\documentclass{article}

\usepackage[preprint,nonatbib]{neurips_data_2024}

\usepackage[utf8]{inputenc} % allow utf-8 input
\usepackage[T1]{fontenc}    % use 8-bit T1 fonts
\usepackage{hyperref}       % hyperlinks
\usepackage{url}            % simple URL typesetting
\usepackage{booktabs}       % professional-quality tables
\usepackage{amsfonts}       % blackboard math symbols
\usepackage{nicefrac}       % compact symbols for 1/2, etc.
\usepackage{microtype}      % microtypography
\usepackage{xcolor}         % colors

\usepackage{amsmath}
\usepackage{graphicx}
\usepackage{wrapfig}
\usepackage{multirow}

\usepackage{amssymb}% http://ctan.org/pkg/amssymb
\usepackage{pifont}% http://ctan.org/pkg/pifont
\usepackage{xcolor} % Required for color
\usepackage{subcaption}

\newcommand{\cmark}{\textcolor[rgb]{0.0, 0.5, 0.0}{\ding{51}}}%
\newcommand{\xmark}{\textcolor{red}{\ding{55}}}%

\newcommand{\Ab}[1]{$\mathbf{X}_{\mathrm{#1}}$}
\newcommand{\Xy}[2]{$\mathbf{#1}_{\mathrm{#2}}$}
\newcommand{\Xyz}[2]{\mathbf{#1}_{\mathrm{#2}}}

\title{LLaVA-Surg: Towards Multimodal Surgical Assistant via Structured Surgical Video Learning}

\author{
  Jiajie Li\thanks{Equal contribution. $^\dag$ Equal advising.}\ \ , Garrett Skinner$^*$, Gene Yang, Brian Quaranto, Steven Schwaitzberg \\
  \textbf{Peter Kim$^\dag$, Jinjun Xiong$^\dag$} \\
  University at Buffalo 
}

\begin{document}

\maketitle

\begin{abstract}
Multimodal large language models (LLMs) have achieved notable success across various domains, while research in the medical field has largely focused on unimodal images. Meanwhile, current general-domain multimodal models for videos still lack the capabilities to understand and engage in conversations about surgical videos. One major contributing factor is the absence of datasets in the surgical field.
In this paper, we create a new dataset, Surg-QA, consisting of 102,000 surgical video-instruction pairs, the largest of its kind so far.
To build such a dataset, we propose a novel two-stage question-answer generation pipeline with LLM to learn surgical knowledge in a structured manner from the publicly available surgical lecture videos. 
The pipeline breaks down the generation process into two stages to significantly reduce the task complexity, allowing us to use a more affordable, locally deployed open-source LLM than the premium paid LLM services. It also mitigates the risk of LLM hallucinations during question-answer generation, thereby enhancing the overall quality of the generated data.
We further train LLaVA-Surg, a novel vision-language conversational assistant capable of answering open-ended questions about surgical videos, on this Surg-QA dataset, and conduct comprehensive evaluations on zero-shot surgical video question-answering tasks. We show that LLaVA-Surg significantly outperforms all previous general-domain models, demonstrating exceptional multimodal conversational skills in answering open-ended questions about surgical videos.
We will release our code, model, and the instruction-tuning dataset.

\end{abstract}

\section{Introduction}

% explain what llava-surg can do
%Surgery, as a multimodal discipline within the medical field, diverges significantly from general medical diagnostics 
Surgery, as a discipline with rich multimodal information in the medical field, diverges significantly from general medical diagnoses
that often depend on static imagery, such as magnetic resonance imaging and chest X-ray. %  \GS{cross-sectional imaging}.
The dynamic nature of surgical procedures with
% \GS{progressive disruption and alteration of anatomy,}
complex sequence of actions and multi-stage processes, cannot be fully captured or understood through a single image.

% Surgery, \GS{and surgical video} diverges significantly from general medical diagnostics that often depend on static imagery, such as magnetic resonance imaging and X-ray \GS{imaging}. %  \GS{cross-sectional imaging}.
% The dynamic nature of surgical procedures involves
% \GS{progressive disruption and alteration of anatomy, that must be performed and interpreted with years of technical experience while balancing complex sequences of actions and multi-stage processes, and cannot be fully captured or understood through a single image.}

% current biomeidcal llm doesn't understand surgical videos, can only hand image modality
The medical field has recently witnessed the significant impact of the Large Language Model (LLM), especially in the arena of medical question answering. Domain-specific LLMs like LLaVA-Med~\cite{li2023llavamed} and Med-PaLM~\cite{medpalmsinghal2022large}, fused with publicly accessible medical question-answer data such as PubMed~\cite{pubmedzhang2023biomedclip}, can assist with inquiries about a biomedical image and meet the safety-critical demands of the medical domain. 
Moreover, general purpose LLMs such as GPT~\cite{openai2024gpt4}, despite not being explicitly aligned to the medical field, have shown great potential and versatility when applied to some specific clinical knowledge areas.
%Meanwhile, the utilization of general purpose LLMs such as GPT~\cite{openai2024gpt4} to assess their performance in specific clinical knowledge areas, despite not being explicitly aligned to the medical field, underscores the potential and versatility of LLMs across diverse healthcare applications. 
However, these models are still limited to processing single images,
% \GS{with superficial, non-technical understanding}
thus falling short of venturing into the surgical domain where the video modality plays a crucial role. 

% lack of data
%Parallel video-text data has proven to be useful for generative pretraining during the self-supervised vision-language modeling, 
The availability of parallel video-text datasets has proven to be useful for  pretraining generative model in a self-supervised manner, 
as demonstrated by conversational multimodal LLMs such as Video-ChatGPT~\cite{maaz2023videochatgpt} and Video-LLaVA~\cite{lin2023videollava}, and text-to-video generative models such as Sora~\cite{soraopenai2024videoworldsimulators}. 
However, obtaining surgical video-text pairs is  more challenging than biomedical image-text pairs or general-domain video-text pairs due to the need of more expensive surgical expertise.
%The difficulty in accessing surgical video-text pairs limits models like LLaVA-Med to unimodal biomedical images, preventing their expansion into the surgical domain.
%However, obtaining surgical video-text pairs proves to be more challenging compared to biomedical image-text pairs or general-domain video-text pairs. The difficulty in accessing surgical video-text pairs limits models like LLaVA-Med to unimodal biomedical images, preventing their expansion into the surgical domain.

% Our work
In this work, we introduce the \textbf{L}arge \textbf{L}anguage and \textbf{V}ision \textbf{A}ssistant for \textbf{Surg}ery (LLaVA-Surg), the first attempt at a surgical multimodal conversational assistant. LLaVA-Surg leverages an adapted LLM that integrates the visual encoder of CLIP~\cite{radford2021learningclip} with Llama~\cite{touvron2023llama} as a language backbone, fine-tuned on generated instructional image-text pairs. Our approach further adapts the design for spatiotemporal video modeling and finetunes the model on video-instruction data to capture temporal dynamics and frame-to-frame consistency relationships available in video data.

A fundamental contribution of this work is the introduction of a novel two-stage question-answer generation pipeline. This pipeline extracts surgical knowledge from widely available surgical lecture videos, resulting in the creation of Surg-QA, a dataset comprising over 102K surgical video-instruction pairs. Each pair consists of a video and its corresponding instructional content in a question-answer format. This extensive and diverse dataset enables LLaVA-Surg's to understand surgical videos and engage in comprehensive conversations about surgical videos.

The major contributions of our paper are as follows:
\begin{enumerate}
    \item \textit{Surg-QA.} We introduce Surg-QA, to the best of our knowledge, the first large-scale surgical video instruction-tuning
    % \JX{not consistent - instruction-following or instruction-tuning?}
    dataset, featuring over 102K surgical video question-answer pairs derived from more than 44K surgical video clips across 2,201 surgical procedures. We also introduce the novel two-step question-answer generation pipeline behind Surg-QA. This pipeline effectively mitigates the issue of LLM hallucination, providing a cost-effective solution for large-scale question-answer generation.
    \item \textit{LLaVA-Surg.} We present LLaVA-Surg, to the best of our knowledge, the first video conversation model capable of expert-level understanding of surgical videos and answering open-ended questions about surgical videos. LLaVA-Surg is trained in under 6 hours using eight A100 GPUs, by fine-tuning a general-domain vision-language model on Surg-QA.  Comprehensive evaluations show that LLaVA-Surg excels in zero-shot surgical video question-answering tasks, outperforming previous models and demonstrating strong multimodal conversational skills.

    % \item \textit{Quantitative video conversation evaluation benchmarking framework.} We propose an automated evaluation framework that, with the assistance of LLMs, examines models' capabilities in surgical video observation and reasoning from multiple perspectives. By comparing the results of our evaluation with expert assessments, we validate the effectiveness of our evaluation framework. Additionally, our model outperforms other models in this domain.
    \item \textit{Open-source.} We will publicly release the surgical video instruction-tuning
    % \JX{not consistent - instruction-following or instruction-tuning?}
    dataset, model, and code for data generation and training to advance research in the surgical domain.
\end{enumerate}
\section{Related Work}

%\paragraph{Surgical Visual Question Answering (Surgical VQA)} Models that 
\textbf{Surgical Video Question Answering (Surgical VQA)} models 
can answer questions based on surgical videos and offer assistance to practicing surgeons and surgical trainees. Early surgical VQA methods were largely discriminative~\cite{twinanda2016endonet,czempiel2020tecnophaserecog,yengera2018lessendo2n2}, treating the task as a classification problem where answers were chosen from a predefined set. They excelled in identifying surgical steps, instruments, and organs, but were limited to closed-set predictions and struggled with 
open-ended questions and answers.
%custom answer sets
% \GS{required to convey the depth of expert-level reasoning within surgical video}.
Recent developments have shifted towards generative methods~\cite{seenivasan2022surgicalvqa, bai2023surgicalvlqa, seenivasan2023surgicalgpt} that produce free-form text sequences but are limited to single-turn conversations, preventing them from engaging in a dialogue or answering follow-up questions. Unlike these models, our LLaVA-Surg model can engage in meaningful multi-turn dialogues, answering surgical questions and providing comprehensive surgical knowledge for an interactive learning experience.

%\paragraph{Multimodal LLM for Biomedical Image Conversations} The integration of multimodal LLM into biomedical image conversations 
\textbf{Multimodal LLM for Biomedical Image Conversations} 
represents a significant advancement in the field of medical artificial intelligence. These models combine text and image understanding to enable more nuanced and contextually aware interactions between clinicians and AI systems. For instance, the LLaVA-Med model demonstrates the potential of multimodal LLMs to interpret and generate detailed medical image descriptions, thereby aiding both diagnostics and patient communication~\cite{li2023llavamed}. The application of such models extends to various tasks including VQA, where they provide accurate and relevant answers based on medical images and related queries~\cite{zhang2023pmcvqa,pal2023medhalt}. This multimodal approach also enhances the ability to perform complex reasoning and decision-making processes, which are critical in clinical settings~\cite{liu2024improved}.
% Advancements in this domain are driven by the integration of large-scale datasets and sophisticated training techniques that align textual and visual information effectively, thus improving the performance and reliability of these models~ .
 Collectively, these developments underscore the transformative potential of multimodal LLMs in enhancing biomedical image conversations and ultimately improving patient care outcomes~\cite{he2020pathological, lau2018dataset}.

%\paragraph{Multimodal LLM for Video Conversations} Multimodal LLMs are transforming video conversations by seamlessly integrating text, images, and video data. Early works like FrozenBiLM~\cite{yang2022zeroshotfrozenbilm} demonstrated the promise of aligning vision and language models for multimodal understanding. Recent advancements like Video-LLaVA~\cite{lin2023videollava}, Video-ChatGPT~\cite{maaz2023videochatgpt}, and ChatUniVi~\cite{jin2024chatunivi} illustrate practical applications in video contexts, delivering real-time, contextually aware responses that improve user interactions. Specifically, Video-LLaVA integrates visual and language data using the Language-Bind framework, enhancing video understanding and generating coherent, contextually relevant responses. Video-ChatGPT excels in handling complex video data, providing detailed analysis and responses. ChatUniVi pushes the boundaries further by integrating unified video and language processing capabilities, facilitating more natural and interactive video conversations. The success of these models in general domains also demonstrates the potential of multimodal models in specialized video fields such as surgery.

\textbf{Multimodal LLM for Video Conversations} has demonstrated great potential by integrating general-domain text, images, and video data. Early works like FrozenBiLM~\cite{yang2022zeroshotfrozenbilm} demonstrates the promise of aligning vision and language models for multimodal understanding. Recent advancements like Video-LLaVA~\cite{lin2023videollava}, Video-ChatGPT~\cite{maaz2023videochatgpt}, and ChatUniVi~\cite{jin2024chatunivi} illustrate practical applications in video contexts, delivering real-time, contextually aware responses that improve user interactions. Specifically, Video-LLaVA integrates visual and language data using the Language-Bind framework, enhancing video understanding and generating coherent, contextually relevant responses. Video-ChatGPT excels in handling complex video data, providing detailed analysis and responses. ChatUniVi pushes the boundaries further by integrating unified video and language processing capabilities, facilitating more natural and interactive video conversations. 
But their applicability to domain-specific videos like surgery
videos have not yet been proven.

\section{Surgical Video Instruction-tuning Data Generation}
\label{sec:3}
\begin{figure*}[hb]
  \centering
  \includegraphics[width=0.9\linewidth]{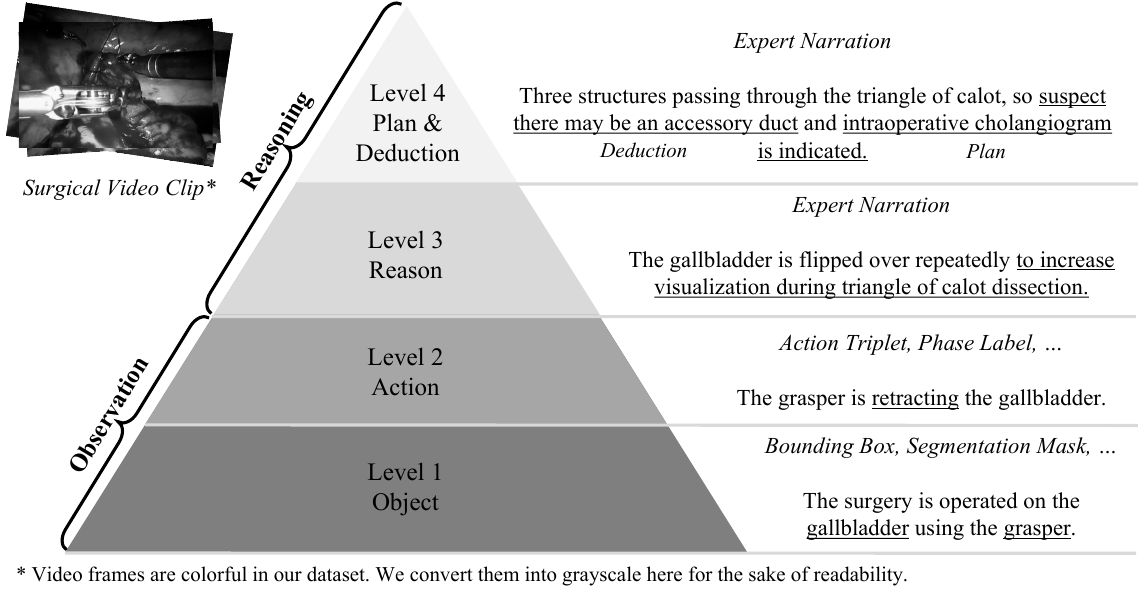}
  %\caption{Surgical Knowledge Pyramid. Surgical video interpretation exists on a spectrum of four levels split into observation and reasoning. Video interpretation at the observation levels represents traditional computer vision paradigms of object detection, segmentation, and labeling but conveys a superficial level of understanding. Interpretation at the reasoning levels provides contextual understanding and reasoning to the observation, conveying deep, surgical expert-level understanding.}
  \caption{Surgical Knowledge Pyramid. Surgical video interpretation can be categorized into four levels. The first two levels represent the observation capabilities, which can be captured by traditional computer vision tasks such as object detection, segmentation, and labeling. But this only conveys a superficial level of understanding. The next two levels represent the reasoning capabilities. Interpretation at the reasoning levels provides the rationale behind the observations, further offering deductions and plannings, conveying deep, surgical expert-level understanding.}
\label{fig:pyramid}
\end{figure*}

There is a significant deficiency in specialized datasets for training multimodal LLM as a conversational assistant in the surgical domain.
As illustrated in Figure~\ref{fig:pyramid}, information in the surgical domain can be categorized into four distinct levels: (1) basic identification of surgical objects such as organs and instruments, (2) recognition of discrete surgical actions, (3) higher-order reasoning of surgical actions, and (4) expert level deduction and planning.

% \footnotetext{This is the footnote text corresponding to the figure caption.}

However, existing datasets~\cite{bai2023surgicalvlqa, yuan2024advancing} lack level 3 and 4 information. To address this, we create \textit{Surg-QA}, the first surgical instruction-tuning dataset that contains all four levels
of information.
%specifically to train the multimodal surgical assistant by mining the widely available surgical lectures through an automated process. 
The proposed dataset consists of 100K video-text pairs from structured learning of surgical lecture videos and 2K pairs focusing on the surgical visual concept alignment.
\begin{figure*}[t]
  \centering
  \includegraphics[width=1.\linewidth]{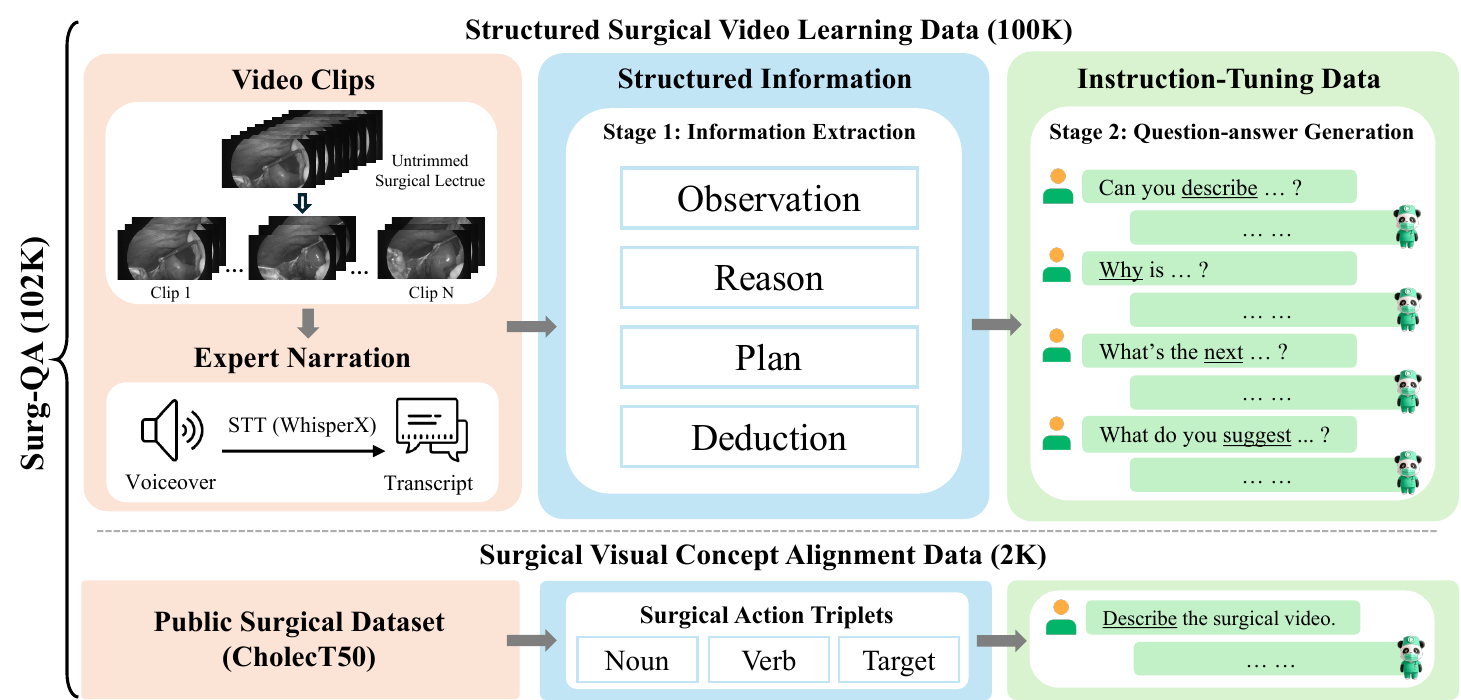}
  \caption{Instruction-Tuning Data Generation Pipeline. Top: Structured surgical video learning begins with untrimmed lecture videos divided into clips. Expert narrations (transcripts) from the lectures are converted to text using WhisperX~\cite{bain2022whisperx}. We then prompt Llama-3-70B to extract the structured information from the transcripts. Finally, the extracted information is provided to Llama-3-70B to generate the instruction-tuning data. Bottom: Surgical visual concept alignment data are concise descriptions of surgical videos, generated based on surgical action triplets.}
  %The progressive levels of surgical video understanding achieved through computer vision and machine learning. }
  \label{fig:data_pipeline}
  \vspace{-0.7cm}
\end{figure*}

\textbf{Surgical Video Instruction-Tuning Data.} For a surgical video \Ab{v} and its transcript \Ab{t}, we prompt Llama-3-70B~\cite{meta2024llama3} through a two-step approach to create a set of questions \Ab{q} that can be answered only when the video is provided, aiming to guide the assistant in describing the video content. A single-round instruction-tuning example can thereby represented by:
\begin{align}
\mathtt{User} :\mathbf{X}_\mathrm{q}\ \mathbf{X}_\mathrm{v} \mathtt{<}\mathtt{STOP}\mathtt{>}
\backslash  \mathrm{n}\
\mathtt{Assistant} :  \mathbf{X}_\mathrm{a} \mathtt{<}\mathtt{STOP}\mathtt{>}\backslash  \mathrm{n}
\label{eq:instruct}
\end{align}
\textbf{Structured Surgical Video Learning.} 
% \GS{To train the model on level 3 and 4 surgical video reasoning,}
We propose a two-step \textit{extraction-generation} approach utilizing the Llama-3-70B model for processing surgical video lectures, as illustrated in Figure~\ref{fig:data_pipeline}.
Specifically, given a surgical lecture video \Ab{v} with voiceover, we begin by applying WhisperX~\cite{bain2022whisperx} to transcribe the spoken content of surgical lecture videos into text. Following this, unlike previous work~\cite{textannogilardi2023chatgpt, liu2024visual, li2023llavamed} that directly prompt LLM to generate multi-round questions and answers based on the text information, we first prompt LLM to extract the key information from the transcripts in a structured manner, focusing on four main components: the observation \Xy{I}{o} and the corresponding reason \Xy{I}{r}, plan \Xy{I}{p} and deduction \Xy{I}{d} as shown in Figure~\ref{fig:pyramid}.
This structured representation of videos ensures high-quality data by extracting only surgery-related information, thus mitigating noise from non-surgical clips or non-informative conversations. Additionally, it reduces the risk of LLM hallucination~\cite{huang2023survey, li2023llavamed} by restricting the model to information extraction only.
We also manually curate few-shot examples to teach how to extract high-quality information based on the transcript. See Appendix~\ref{app:prompt} for the prompt and few-shot examples.

\begin{figure*}[t]
\centering
  \begin{minipage}[b]{0.95\linewidth}
    \centering
    \begin{subfigure}[b]{\linewidth}
      \centering
      \includegraphics[width=\linewidth]{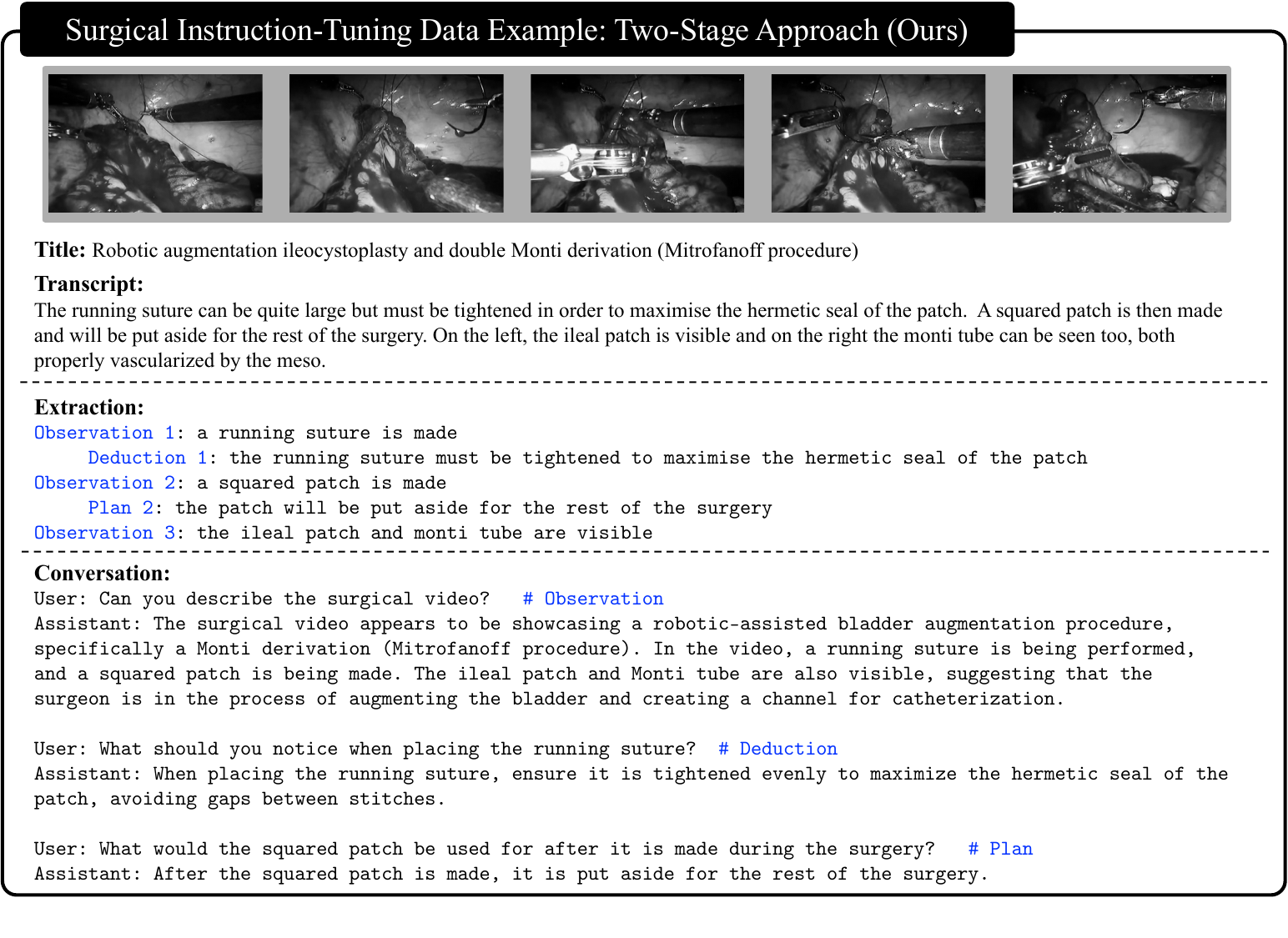}
    \vspace{-0.7cm}
      % \caption{An instance of our instruction-tuning data. Top:  Frames of a surgical lecture video, with the video title and transcript. Middle: The structured information extracted by Llama-3-70B from the title and transcript. Bottom: The instruction-tuning data generated by Llama-3-70B using structured information.}
      % \label{fig:ifexample}
    \end{subfigure}
    \begin{subfigure}[b]{\linewidth}
      \centering
      \includegraphics[width=\linewidth]{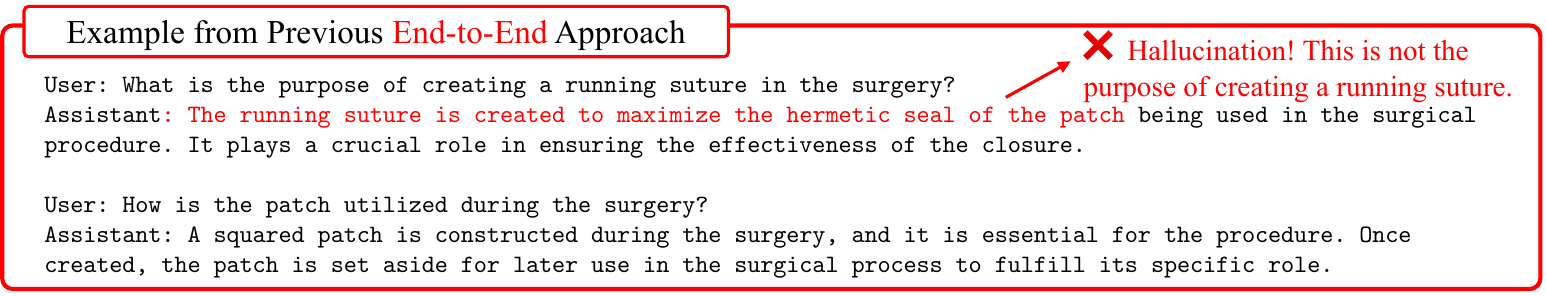}
      % \caption{Question-answer pairs generated with the same information as Figure~\ref{fig:ifexample} using the End-to-end approach introduced in previous works such as LLaVA and LLaVA-Med.}
      % \label{fig:badexample}
    \end{subfigure}
  \caption{Comparison of instruction-tuning data generated by our two-stage approach (top) and the previous end-to-end approach (bottom). Both approaches were given the same video title and transcript. Our approach accurately extracted information from the transcript, generating correct question-answer pairs. In contrast, the conventional end-to-end approach produced incorrect question-answer pairs due to hallucination.}
\label{fig:combinedexample}
  \end{minipage}
  \vspace{-0.3cm}
  % Top:  Frames of a surgical lecture video, with the video title and transcript. Middle: The structured information extracted by Llama-3-70B from the title and transcript. Bottom: The instruction-tuning data generated by Llama-3-70B using structured information.}
  \vspace{-0.3cm}
\end{figure*}

Once the information has been extracted, we can create the instruction-tuning data as multi-turn conversations by prompting LLM to generate different types of question-answering pairs in a controllable way. For example, by concatenating all the observations $(\Xyz{I}{o}^1, \Xyz{I}{o}^2,\dots, \Xyz{I}{o}^T)$ where $T$ is the total observations of \Ab{v}, we prompt LLM to generate the first question-answer pair $[\Xyz{X}{q}^1, \Xyz{X}{a}^1]$ that focus on the visual content of the surgical lecture. Next, for each of the $[\Xyz{I}{o}, \Xyz{I}{r}]$, $[\Xyz{I}{o}, \Xyz{I}{p}]$ and $[\Xyz{I}{o}, \Xyz{I}{d}]$ combinations, we prompt LLM to generate the surgical reasoning question-answering pairs $(\Xyz{X}{q}^2, \Xyz{X}{a}^2,\dots, \Xyz{X}{q}^N, \Xyz{X}{a}^N)$ where $N$ is the total number of question-answer pairs. By stacking the question-answer pairs, we can create a multi-turn conversation, where the instruction $\mathbf{X}_\mathtt{q}^{t}$ at the $t$-th turn is defined as:
\begin{align}
    \mathbf{X}^{t}_{\mathtt{q}} = 
\begin{cases} 
[\mathbf{X}^{1}_{q}, \mathbf{X}_{v}] \text{ or } [\mathbf{X}_{v}, \mathbf{X}^{1}_{q}], & t = 1 \\
\mathbf{X}^{t}_{q}, & t > 1 
\end{cases}
\end{align}
% For each video clip X, we generate multi-turn conversation data (X1 q , X1 a , · · · , XT q , XT a ), where T is the total number of turns. We organize them as a sequence, by treating all answers as the assistant’s response, and the instruction Xt instruct at the t-th turn as:
We can then construct the multi-turn multimodal instruction-tuning data:
\begin{equation}
\begin{split}
\mathtt{User} :\mathbf{X}_\mathrm{q}^1\ \mathbf{X}_\mathrm{v} \mathtt{<}\mathtt{STOP}\mathtt{>} \backslash  \mathrm{n}\
\mathtt{Assistant} :  \mathbf{X}_\mathrm{a}^1 \mathtt{<}\mathtt{STOP}\mathtt{>}\backslash  \mathrm{n} \\
\mathtt{User} :\mathbf{X}_\mathrm{q}^2\ \mathtt{<}\mathtt{STOP}\mathtt{>} \backslash  \mathrm{n}\
\mathtt{Assistant} :  \mathbf{X}_\mathrm{a}^2 \mathtt{<}\mathtt{STOP}\mathtt{>}\backslash  \mathrm{n} \dots\dots
\label{eq:instruct-multi}
\end{split}
\end{equation}
An example of instruction-tuning data is shown in Figure~\ref{fig:combinedexample}. In comparison, we provide the pairs generated with the same information using the previous end-to-end approach~\cite{li2023llavamed, liu2024visual}, the previous approach generated an incorrect pair due to the hallucination. The prompt for structured information extraction is provided in Appendix~\ref{app:prompt}.

\begin{figure*}[t]
  \centering
  \begin{minipage}[b]{0.33\linewidth}
    \centering
    \begin{subfigure}[b]{\linewidth}
      \centering
      \includegraphics[width=\linewidth]{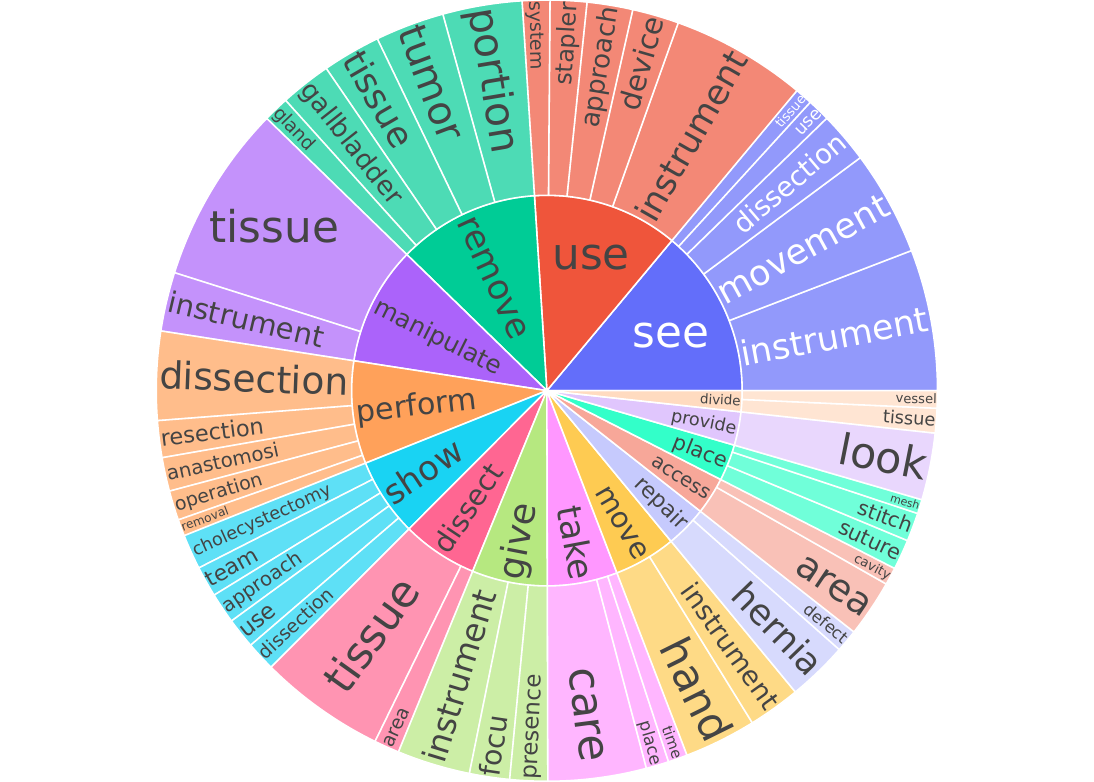}
      \caption{Observation}
    \end{subfigure}
    \vfill
    \begin{subfigure}[b]{\linewidth}
      \centering
      \includegraphics[width=\linewidth]{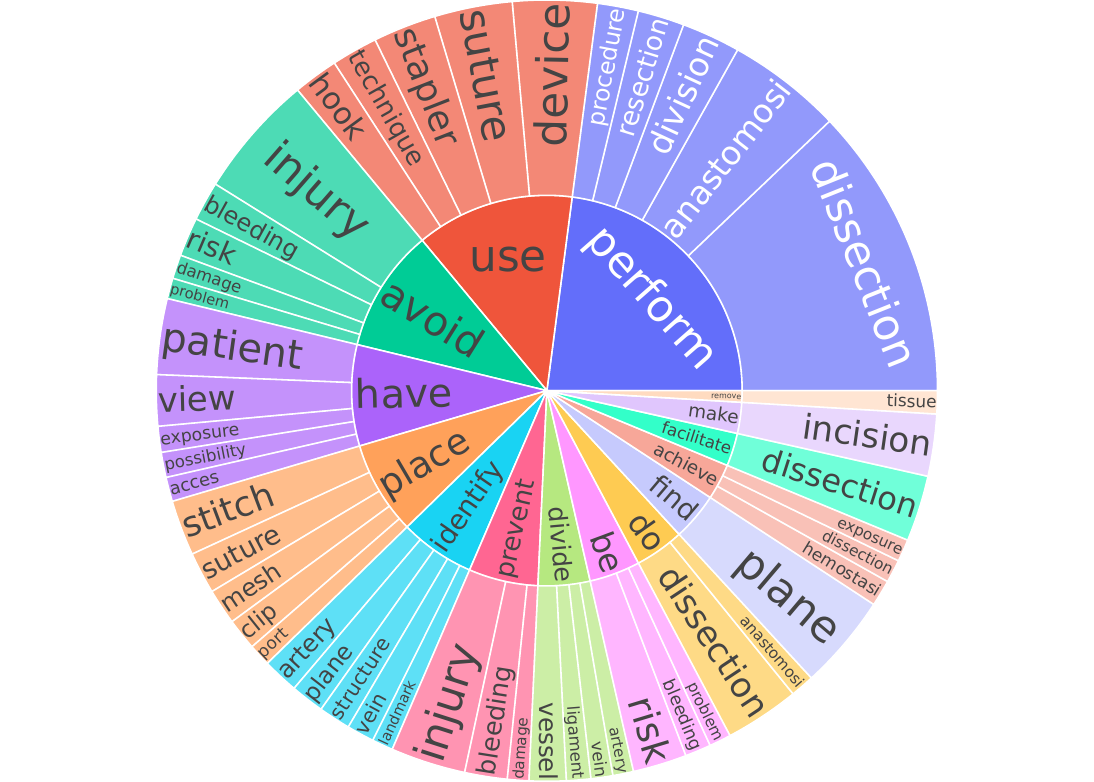}
      \caption{Reasoning}
    \end{subfigure}
  \end{minipage}
  \hfill
  \begin{minipage}[b]{0.66\linewidth}
      \centering
    \begin{subfigure}[b]{\linewidth}
      \centering
      \includegraphics[width=\linewidth]{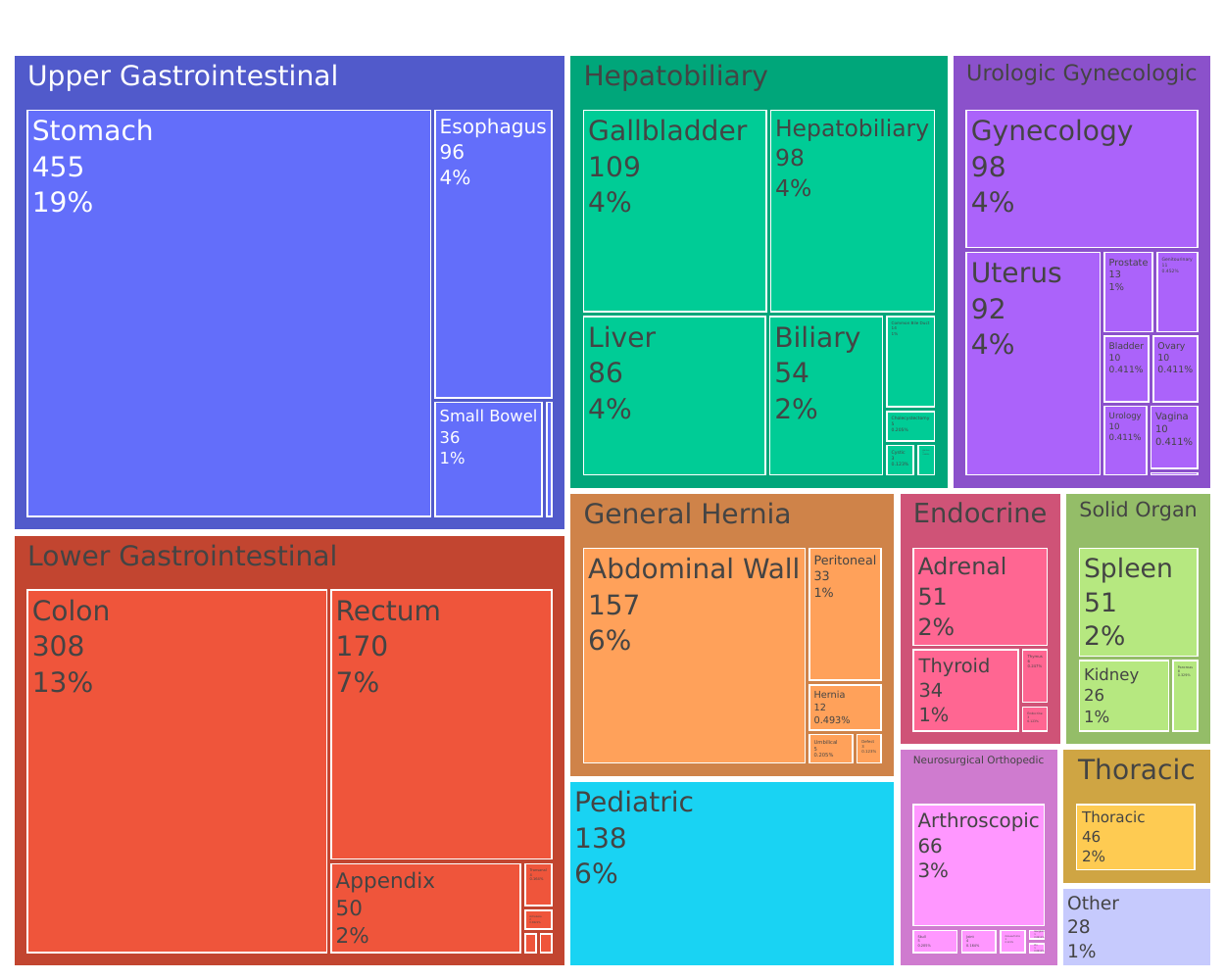}
      \caption{Surg-QA surgery instance category treemap.}
      \label{fig:treemap}
    \end{subfigure}
  \end{minipage}
% \vspace{0.5cm}
\begin{minipage}[b]{\linewidth}
\centering
\begin{subfigure}[b]{\linewidth}
  \centering
  \includegraphics[width=\linewidth]{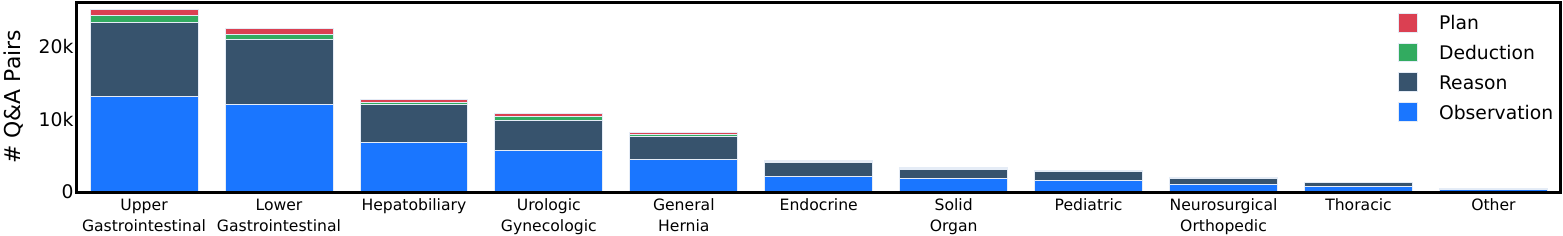}
  \caption{Frequencies of QA pairs by categories.}
\end{subfigure}
\end{minipage}
  \caption{The data statistics of surgical multimodal instruction-tuning data: (a,b) The root verb-noun pairs provide an overview of our dataset of instructions and responses. In the plot, the inner circle represents the root verb of the response, and the outer circle represents the direct nouns. (c) The distribution of videos of different types. (d) The distribution of video and QA pairs on 11 categories.}
  \label{fig:surg_qa_stat}
  \vspace{-0.5cm}
\end{figure*}

We collected 2,151 surgical lecture videos from WebSurg\footnote{\url{https://www.websurg.com}}~\cite{websurg2024}. As shown in Figure~\ref{fig:treemap}, these videos cover upper and lower gastrointestinal, hepatobiliary, urologic, gynecologic, general hernia, pediatric, endocrine, solid organ, and thoracic surgeries. We divided them into 42K short clips (15-30 seconds). Our automated pipeline generated 100K video-text pairs.
We provided detailed statistics of Surg-QA in Figure~\ref{fig:surg_qa_stat}.

% \input{figures/instruction-tuning-example}
% \input{figures/bad_example}
% \begin{figure*}[t]
% \centering
%   \begin{minipage}[b]{0.95\linewidth}
%     \centering
%     \includegraphics[width=\linewidth]{assets/eg_new.pdf}
%     \vspace{-0.9cm}
%     \caption{An instance of our instruction-tuning data. Top:  Frames of a surgical lecture video, with the video title and transcript. Middle: The structured information extracted by Llama-3-70B from the title and transcript. Bottom: The instruction-tuning data generated by Llama-3-70B using structured information.}
%     \label{fig:ifexample}
%   \end{minipage}
  
%   \centering
%   \begin{minipage}[b]{0.95\linewidth}
%     \centering
%     \includegraphics[width=\linewidth]{assets/eg_bad.pdf}
%     \vspace{-0.5cm}
%     \caption{Question-answer pairs generated with the same information as Figure~\ref{fig:ifexample} using the End-to-end approach introduced in previous works such as LLaVA and LLaVA-Med.}
%     \label{fig:badexample}
%   \end{minipage}
% \vspace{-0.5cm}
% \end{figure*}

\textbf{Surgical Visual Concept Alignment.}
We create the surgical visual concept alignment data based on the public surgical dataset CholecT50, which aids the model in recognizing fundamental surgical visual concepts such as instruments, organs, and actions.
% We the public surgical dataset CholecT50~\cite{nwoye2021rendezvous} for surgical visual concept alignment which helps the model recognize the basic surgical visual concepts such as instruments and organs.
CholecT50 includes 50 endoscopic videos, each frame annotated with action triplets: $[\mathtt{instrument}, \mathtt{verb}, \mathtt{target}]$ that denote the tool, action, and the object or site of the action, respectively. We first divide the videos into 30-60-second clips. To generate a concise description for each video clip, we begin by merging consecutive frames with the same annotations while preserving the chronological order. Once this sequence of merged annotations is obtained, we use the sequence to prompt a Llama-3-70B to generate a description of the clip. In total, we sampled 2,200 video-text pairs to create the instruction-tuning dataset as outlined in Equation~\ref{eq:instruct}. 

\textbf{Comparisons.} 
We compare Surg-QA with both existing general-domain VQA datasets and surgical-domain VQA datasets as shown in Tables~\ref{tab:compare_general} and \ref{tab:compare_surgical}.
% Firstly, in addressing the question of whether Surg-QA is enough to train a multimodal LLM. As shown in Table~\ref{tab:compare_general}, even in comparison to the general-domain VQA dataset, our Surg-QA dataset proves to be substantial in size. It contains 44K videos and 102K QA pairs, making it comparable with other general domain instruction-tuning datasets.
First, regarding whether Surg-QA is sufficient to train a multimodal LLM: Table~\ref{tab:compare_general} demonstrates that Surg-QA is substantial in size, with 44K videos and 102K QA pairs, making it comparable to general-domain VQA datasets.
Second, Surg-QA surpasses traditional surgical-domain VQA datasets. As shown in Table~\ref{tab:compare_surgical}, Surg-QA includes more surgical procedures, and a wider range of surgical types (Figure~\ref{fig:treemap}), and provides video-wise question-answer pairs rather than frame-wise annotations. It also integrates both observational and reasoning-based knowledge, offering a comprehensive understanding of surgical procedures.

\begin{table}[t]
\centering
\begin{minipage}{\linewidth}
  \centering
    % \begin{table}[h]
%     \label{tab:comp_general}
%     \centering
\resizebox{.9\textwidth}{!}{
    \begin{tabular}{c|cccc}
    \toprule
    \textbf{General VQA Datasets}        & \textbf{Q\&A pairs generation} & \textbf{\# Video clips} & \textbf{\# Q\&A pairs} & \textbf{Avg. length} \\
    \midrule
    MSVD-QA~\cite{xu2017video}  & Automatic & 2K & 51K & 10s \\
    ActivityNet-QA~\cite{yu2019activitynetqa}  & Human & 6K & 60K & 180s \\
    MovieQA~\cite{tapaswi2016movieqa}  & Human & 7K & 7K & 200s \\ 
    MSRVTT-QA~\cite{xu2017video} & Automatic &  10K & 244K & 15s \\
    % TGIF-QA & Automatic \& Human & OE \& MC & 56,720 & 103,919 & 3s \\ 
    % Video-QA  & Automatic & 18K & 175K & 45s \\
    VideoInstruct-100K~\cite{maaz2023videochatgpt} & Human\&Automatic & --  & 100K & - \\
    \midrule
    Surg-QA (Ours) & Automatic & 44K & 102K & 20s \\
    \bottomrule
    \end{tabular}
    % }
    % \begin{tabular}{l|r|r|l}
    % \toprule
    % Datasets & \# Videos & \# QA pairs & Avg. length \\
    % \midrule
    % MSVD-QA & 1,970 & 50,505 & 10s \\
    % MSRVTT-QA & 10,000 & 243,680 & 15s \\
    % TGIF-QA & 56,720 & 103,919 & 3s \\ 
    % MovieQA & 6,771 & 6,462 & 200s \\ 
    % Video-QA & 18,100 & 174,775 & 45s \\
    % ActivityNet-QA & 5,800 & 58,000 & 180s \\
    % \midrule
    % Surg-QA (Ours) & 42,000 & 126,000 & 22s \\
    % \bottomrule
    % \end{tabular}
    
    }
%     \vspace{0.1cm}
% \caption{Comparison of existing VideoQA datasets with ours (OE: open-ended, and MC: multiple-choice).}
% % TODO comparison with other Medical Datasets

% \end{table}
    \vspace{0.1cm}
  \caption{Comparison with existing general-domain VQA datasets.}
  \label{tab:compare_general}
\end{minipage}%
\\
\vspace{-0.1cm}
\begin{minipage}{\linewidth}
  \centering
    % \begin{table}[h]
%     \label{tab:comp_surg}
%     \centering
\resizebox{\textwidth}{!}{
    \begin{tabular}{c|ccccc}
    \toprule
    \multirow{2}{*}{\textbf{Surgical VQA Dataset}} & \multirow{2}{*}{\textbf{\# Surgical procedures}} & \multirow{1}{*}{\textbf{Total length}}& \multirow{2}{*}{\textbf{Video-wise Q\&A}} & \multicolumn{2}{c}{\textbf{Knowledge}} \\ 
    & &\multicolumn{1}{c}{\textbf{(Hour)}} & & \multicolumn{1}{c}{\textbf{Observation}} & \textbf{Reasoning} \\ \midrule
    EndoVis-18-VQA~\cite{seenivasan2022surgicalvqa} & 14 & -- & \xmark & \cmark & \xmark \\ 
    Cholec80-VQA~\cite{seenivasan2022surgicalvqa} & 80 & 24&\xmark & \cmark  & \xmark  \\ 
    SSG-VQA~\cite{yuan2024advancing} & 40 & 28 & \xmark & \cmark  & \xmark  \\ 
    \midrule
    Surg-QA (Ours) & 2201 & 233 & \cmark & \cmark  & \cmark \\ \bottomrule
    \end{tabular}
    }
%     \vspace{0.1cm}
% \caption{Comparison of existing surgical VQA datasets with ours.}

% \end{table}
    \vspace{0.1cm}
  \caption{Comparison with existing surgical-domain VQA datasets.}
  \label{tab:compare_surgical}
\end{minipage}
\vspace{-0.7cm}
\end{table}
\section{Surgical Visual Instruction Tuning}
\textbf{Architecture.} LLaVA-Surg is a large vision-language model that aims to generate meaningful conversation about surgical videos. It employs the architecture of Video-ChatGPT~\cite{maaz2023videochatgpt}, a general-domain multimodal conversation model. Given a video, the model first samples $N$ frames uniformly, and calculate the frame-level features $h \in \mathbb{R}^{N \times h \times w \times D}$ for each of the frames using CLIP ViT-L/14~\cite{radford2021learningclip}, where $D$ is the hidden dimension of CLIP features and $h, w$ are the video height and width respectively. The features $h$ are fused through a temporal-fusion operation, where the temporal features $t\in \mathbb{R}^{N \times D}$ are derived through an average-pooling operation along the temporal dimension, and spatial features $s\in \mathbb{R}^{(h\times w) \times D}$ are derived using the same average-pooling operation but along the spatial dimensions. By concatenating $t$ and $s$, we derived the video-level features $f\in \mathbb{R}^{(N+h\times w) \times D}$, then feed it through a linear projection layer that connects $f$ to the language model. %We use the language backbone from pretrained LLaVA-Med as the initial medical-domain language model.

\textbf{End-to-End Instruction-Tuning.}
To balance the knowledge from levels 1 to 4, we combine the structured surgical video learning data and concept alignment data as discussed in Section~\ref{sec:3}, this results in 38K training video clips with 90K question-answer pairs. These pairs are converted to instruction-following data as described in Equation~\ref{eq:instruct-multi}, the data includes instructions that simply present the task of describing the video, and tasks that answer various reasoning tasks. To train the model to follow various instructions and complete tasks in a conversational manner, we finetune LLaVA-Surg as a chatbot on the conversational data. During our training, we keep the weights of the CLIP visual encoder only and finetune the rest of the parameters.

\section{Experiments}
We conduct experiments to study two key components: the performance of LLaVA-Surg and the quality of the produced multimodal surgical instruction-tuning data. Our experiments focus on two evaluation settings: (1) How does LLaVA-Surg perform in surgical video question-answering, and how does it compare to existing methods in the surgical domain? (2) How does the GPT evaluation framework compare to the human expert evaluation?

\subsection{Implementation Details}
\textbf{Data.} We collected 2,054 surgical procedures from WebSurg using the keyword "intervention" and an additional 97 procedures with the keyword "gallbladder" for future evaluation purposes, totaling 2,151 procedures. These were randomly divided into a training set of 1,935 procedures and a test set of 216 procedures.
In our instruction-tuning data generation pipeline, we use the 'large-v2' version of WhisperX~\cite{bain2022whisperx} to transcribe the surgical lectures. We use Llama-3-70B-Instruct~\cite{meta2024llama3} for information extraction and data generation as mentioned in Section~\ref{sec:3}. We use 'gpt-3.5-turbo-0125' to perform the following quantitative evaluation.

\textbf{Training.} We use LLaVA-Med as our pre-trained language backbone and finetune the model on 90K surgical video instruction following data. We use CLIP ViT-L/14 as the image encoder and use LLaVA-Med's language backbone as the initial weight of LLaVA-Surg. We update the linear layer projecting the video features to the LLM's input space and the language backbone, while the CLIP encoder is kept frozen. We finetune the model for 5 epochs using a learning rate of 2e-5 and an overall batch size of 128. The training of our 7B model took around 6 hours on 8 A100 40GB GPUs. For the rest of the hyperparameters, we follow the settings in \cite{maaz2023videochatgpt}.

\subsection{Quantitative Evaluation}\label{sec:gpt_eval}
% 1. Comparison with Video-ChatGPT, Video-LLaMA, Gemini Pro
\begin{table}[h]
\centering
\begin{tabular}{l|ccc}
\toprule
Model             & Score (0-5)   & Accuracy@all          & Accuracy@1        \\ \midrule
LLaVA-Med         & 1.30          & 0.123          & 0.211          \\
Video-LLaVA       & 1.32          & 0.129          & 0.224          \\
Video-ChatGPT     & 1.04          & 0.098          & 0.172          \\ \midrule
LLaVA-Surg (Ours) & \textbf{2.45} & \textbf{0.308} & \textbf{0.545} \\ \bottomrule
\end{tabular}
\vspace{0.2cm}
\caption{\textbf{Zeroshot Surigcal Question-Answering} comparison of LLaVA-Surg with other video generative models on test split of Surg-QA.}
\label{tab:gpt_eval}
\vspace{-0.4cm}
\end{table}

\textbf{Question-Answer Evaluation.}
We conducted a comprehensive quantitative evaluation on the test split of Surg-QA consisting of 4359 open-ended surgical video question-answer pairs. Following recent works~\cite{lin2023videollava,maaz2023videochatgpt,li2023llavamed} that use GPT to evaluate open-ended questions, our evaluations employ GPT-3.5-Turbo for evaluation to assess the model’s capabilities of answering surgical video questions. This evaluation process measures the accuracy of the model’s generated predictions and assigns a relative score on a scale from 0 to 5. We provide the prompt used for evaluation in Appendix~\ref{app:prompt}.

% \JX{the following seemed to be repeating some words from the previous paragraph.} We conducted a comprehensive quantitative evaluation on the test split of Surg-QA, which consists of 4,359 open-ended surgical video question-answer pairs. Following recent works~\cite{lin2023videollava,maaz2023videochatgpt,li2023llavamed} that use GPT for evaluating open-ended questions, we employed GPT-3.5-Turbo to assess the model's capabilities in answering surgical video questions.
In our evaluation process, GPT-3.5-Turbo was utilized to score the model's outputs by comparing them with the ground truth from the dataset. Each output was rated on a scale from 0 to 5 based on how accurately it reflected the observations. This approach enables us to directly determine the accuracy of the model's predictions. To achieve this, we provided GPT with the extracted observations as mentioned in Section~\ref{sec:3}, allowing it to evaluate the correctness of the observations included in the answers. Additionally, GPT-3.5-Turbo offered detailed comments highlighting the matches and discrepancies for further reference.
% Specifically, for each of the model's outputs, GPT-3.5-Turbo is provided with the model's output and the ground truth observations from the dataset to assign a relative score on a scale of 0-5 that measures how well the observations and reflected in the model's answer. It also allows the model to directly derive the accuracy of the model's generated predictions. To derive the accuracy, we provided the extracted observations mentioned in Section~\ref{sec:3} to GPT, allowing it to evaluate how many of the observations were correctly mentioned in the answers. We also ask GPT-3.5-Turbo to provide with a detailed comment with what are matches and what are not for reference.
Our results are presented in Table \ref{tab:gpt_eval}, where we provide the GPT evaluation scores. Additionally, we calculated the accuracy when at least one observation is matched (accuracy@1) and the overall accuracy for all observations in the test set (accuracy@all).

To benchmark LLaVA-Surg, we compared its performance with other significant models such as Video-LLaVA and Video-ChatGPT. Despite the solid foundation established by these models, LLaVA-Surg outperformed them in the surgical domain, achieving state-of-the-art (SOTA) performance. We also compare with LLaVA-Med which is an MLLM in the biomedical image domain that supports only unimodal images, we feed the first frame of the video clip into the model, and the results demonstrate the importance of video modality to the surgical domain. These results indicate LLaVA-Surg’s ability to understand the surgical video content and generate accurate, contextually rich answers to questions.

\begin{wrapfigure}{}{0.4\textwidth}
  \centering
  \includegraphics[width=\linewidth]{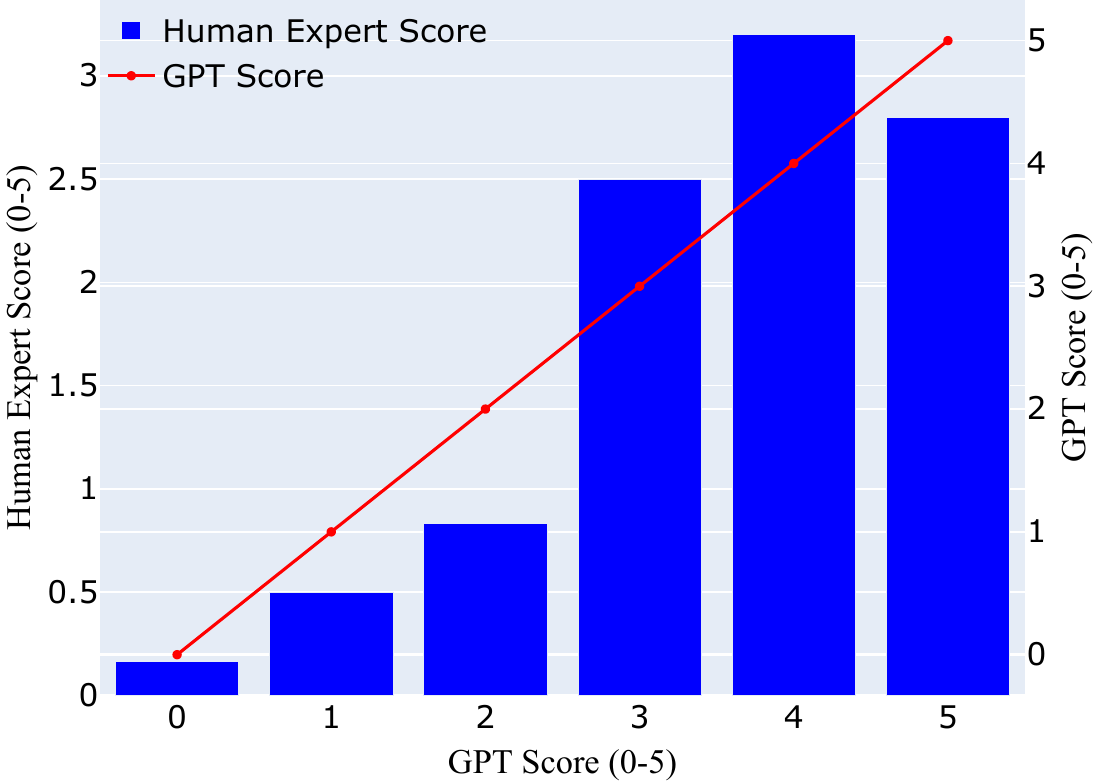}
  \caption{Human Expert vs GPT-3.5-Turbo Evaluation.
  Spearman's rank correlation coefficient $\rho=0.94$.
  }
  \label{fig:human_eval}
  % \vspace{-0.5cm}
\end{wrapfigure}

\textbf{Human Expert Evaluation.}
To validate whether the GPT evaluation framework can benchmark the model's true performance, a human expert (surgeon) also evaluated a subset of the experimental results. The surgeon assigned a score from 0 to 5 to LLaVA-Surg's response based solely on his understanding of the surgical videos. We also provided him with the prompt used for GPT evaluation as a reference. Specifically, we randomly sampled a portion of data for each score from the GPT evaluation results, resulting in a total of 41 video-text pairs, then these samples were compared with the average scores given by the human expert for each score.
We report the comparison results between human expert evaluation and GPT evaluation in Figure \ref{fig:human_eval}. The figure reveals a strong correlation between the scores assigned by the human expert and GPT. This similarity in the scoring patterns indicates that the GPT evaluation framework effectively aligns with human expert judgment and offers a more scalable solution for evaluation. The full human evaluation results with comments are provided in the supplementary materials.

% \paragraph{Perplexity}
\vspace{-0.1cm}
\subsection{Qualitative Evaluation}
\vspace{-0.1cm}
We performed an extensive evaluation of our model on various open-ended surgical video question-answering tasks. Table~\ref{fig:eg_1} illustrates an example involving a gastric bypass procedure. LLaVA-Surg accurately identifies the procedure as a gastric bypass, noting the use of a suture and the closing operation. It correctly answers the subsequent question regarding using a non-absorbable monofilament suture to close the Petersen space. However, LLaVA-Med fails to correctly describe the video, nor answer the following question. We provide more examples in Appendix~\ref{app:more_results}.
\begin{table*}[t]
  \centering
  \includegraphics[width=\linewidth]{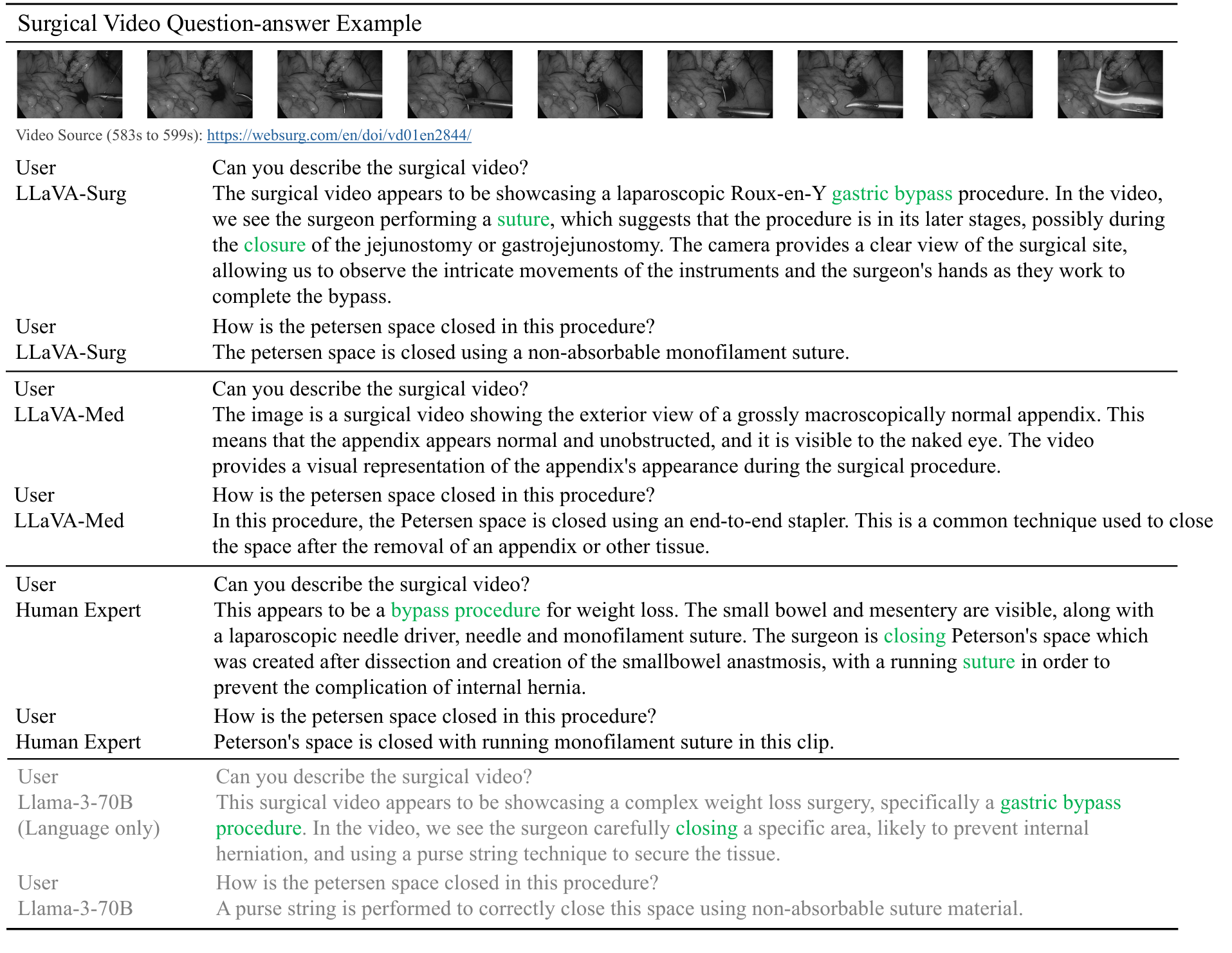}
  \vspace{-0.6cm}
  \caption{Example comparison of surgical video question-answering. We provided the ground truth answers generated by the language-only Llama-3-70B for reference. The answers are based solely on extracted information and the video title. It is considered the model's performance upper bound.}
  \label{fig:eg_1}
  \vspace{-0.5cm}
\end{table*}

\vspace{-0.1cm}
\section{Conclusion}
\label{sec:conclusion}
\vspace{-0.1cm}
In this paper, we introduced Surg-QA, a surgical video instruction-tuning dataset of 102K video-text pairs. Surg-QA is generated primarily through a cost-efficient, two-stage question-answer generation pipeline, which effectively reduces hallucinations during question-answer generation by LLM.
We then trained LLaVA-Surg, a multimodal LLM in the surgical video domain, on Surg-QA. LLaVA-Surg shows great potential in understanding surgical videos and engaging in surgical video conversations, outperforming previous multimodal LLMs in our comprehensive evaluation.
While LLaVA-Surg performs competitively compared to existing methods in the surgical video domain, we note that LLaVA-Surg is limited by hallucinations.
Future work is directed toward engaging experts to review the generated samples in Surg-QA to improve the accuracy and reliability of LLaVA-Surg.
% and weak in general-domain reasoning to many LMMs.
% \JX{In the abstract, we claimed that we removed the hallucination. Please check}.
% , \GS{designed to comprehensively convey levels 1-4 of surgical video reasoning for machine learning}.
% \GS{and investigating novel applications of surgical video LMMs in surgical education and surgical guidance.}

% TODO CHANGE TO OTHERS
\newpage
\bibliographystyle{plain}
\bibliography{references}

\appendix

\section{Data}
\subsection{Surg-QA}
\label{app:data}
We open-source the surgical instruction-tuning dataset Surg-QA following CC BY NC 4.0 license.

\paragraph{Instruction-Tuning Data} See supplementary materials.
\paragraph{Videos} Available in \url{https://websurg.com/}, we provide the corresponding URL to each of the question-answer pair.

\subsection{Prompts}
\label{app:prompt}

\begin{figure*}[h]
  \centering
  \includegraphics[width=\linewidth]{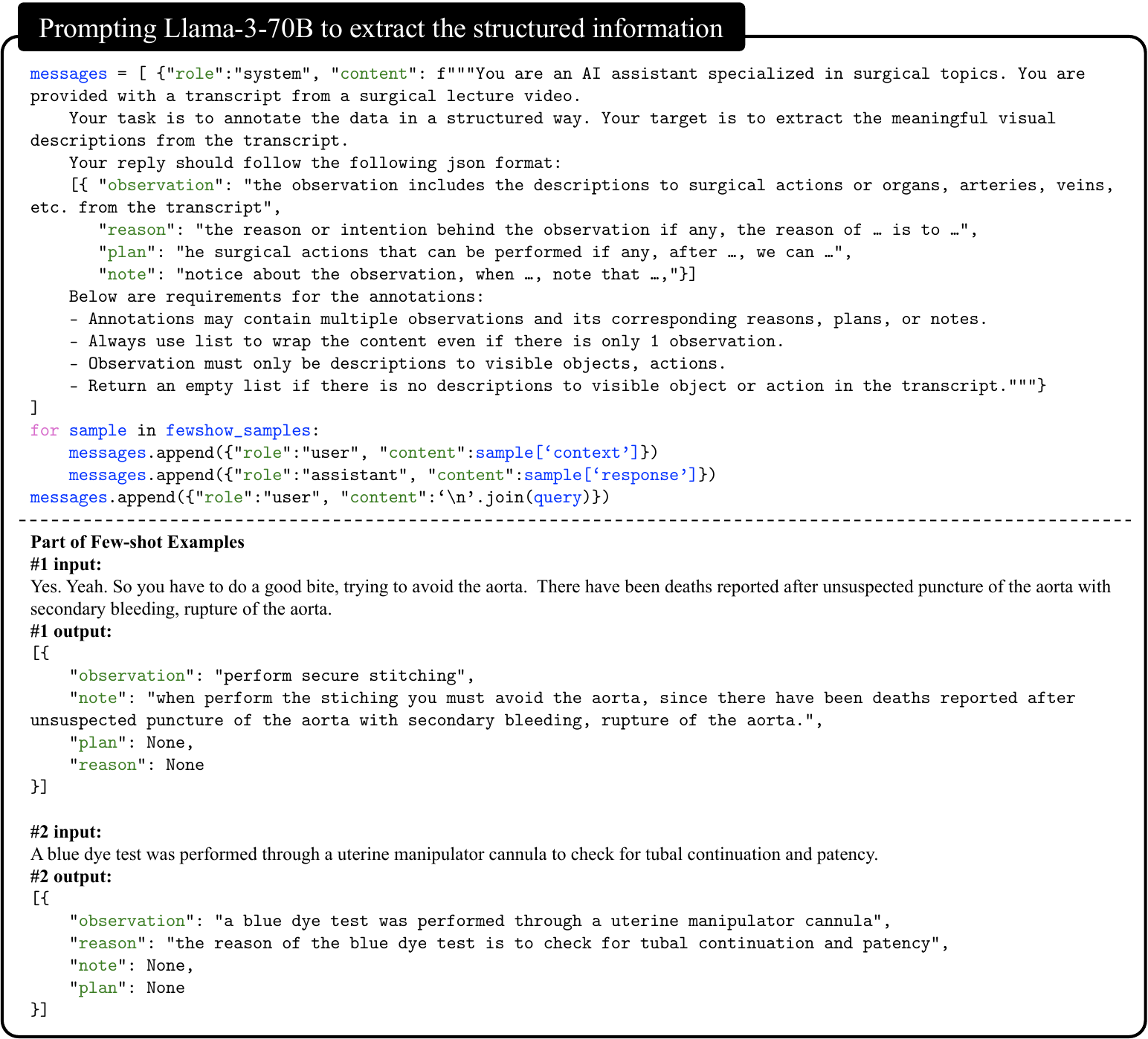}
  \vspace{-0.5cm}
  \caption{\textcolor{blue}{messages} we use to prompt Llama-3-70B to extract structured information. \textcolor{blue}{query} contains the transcribed text for each video clip and the video title.}
  \label{fig:prompt_extract}
  \vspace{-0.5cm}
\end{figure*}

\paragraph{Prompt for information extraction} The prompt used to structurally extract key information from video title and transcript are in Figure~\ref{fig:prompt_extract}.

\begin{figure*}[h]
  \centering
  \includegraphics[width=\linewidth]{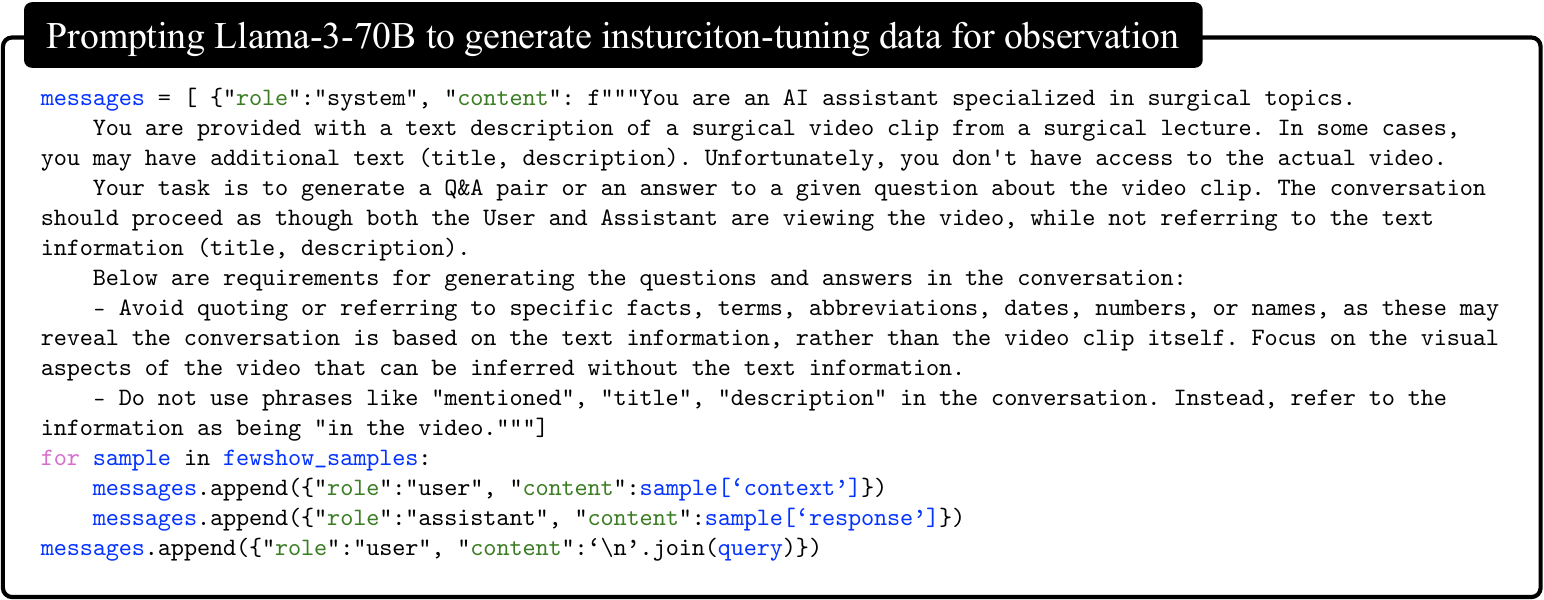}
  \vspace{-0.5cm}
  \caption{\textcolor{blue}{messages} we use to prompt Llama-3-70B to generate instruction-tuning data for observation. \textcolor{blue}{query} contains the concatenated observations.}
  \label{fig:prompt_observation}
  \vspace{-0.5cm}
\end{figure*}

\paragraph{Prompt for question-answer generation for observation} The prompt used to generate instruction data 
that describes a surgical video is in Figure~\ref{fig:prompt_observation}.

\begin{figure*}[h]
  \centering
  \includegraphics[width=\linewidth]{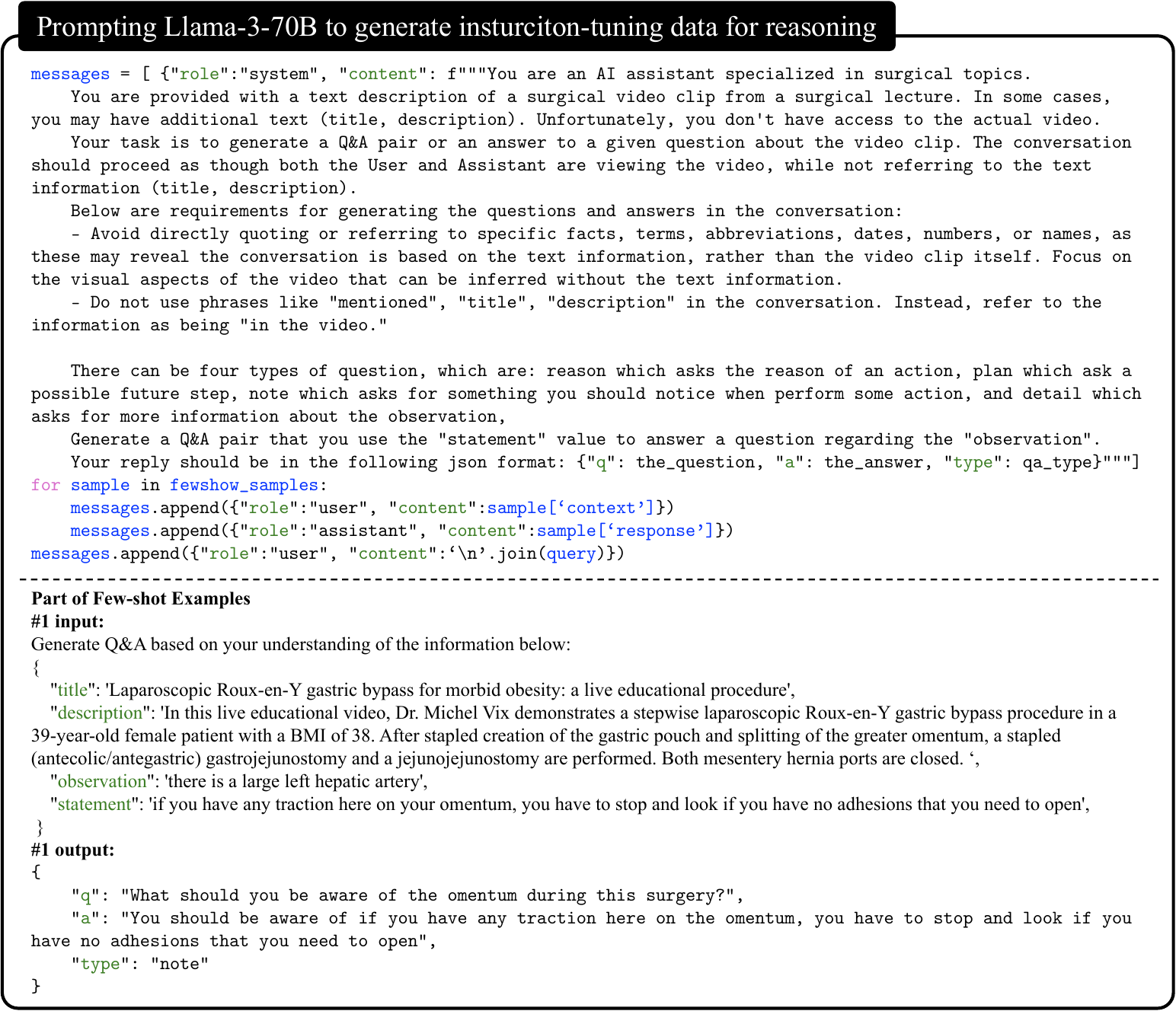}
  \vspace{-0.5cm}
  \caption{\textcolor{blue}{messages} we use to prompt Llama-3-70B to generate instruction-tuning data for reasoning. \textcolor{blue}{query} provides a title, video description, observation, and statement to form a reasoning question-answer pair.}
  \label{fig:prompt_reasoning}
  \vspace{-0.5cm}
\end{figure*}

\paragraph{Prompt for question-answer generation for reasoning} The prompt used to generate instruction data for a variety of reasoning tasks is in Figure~\ref{fig:prompt_reasoning}.

\begin{figure*}[h]
  \centering
  \includegraphics[width=\linewidth]{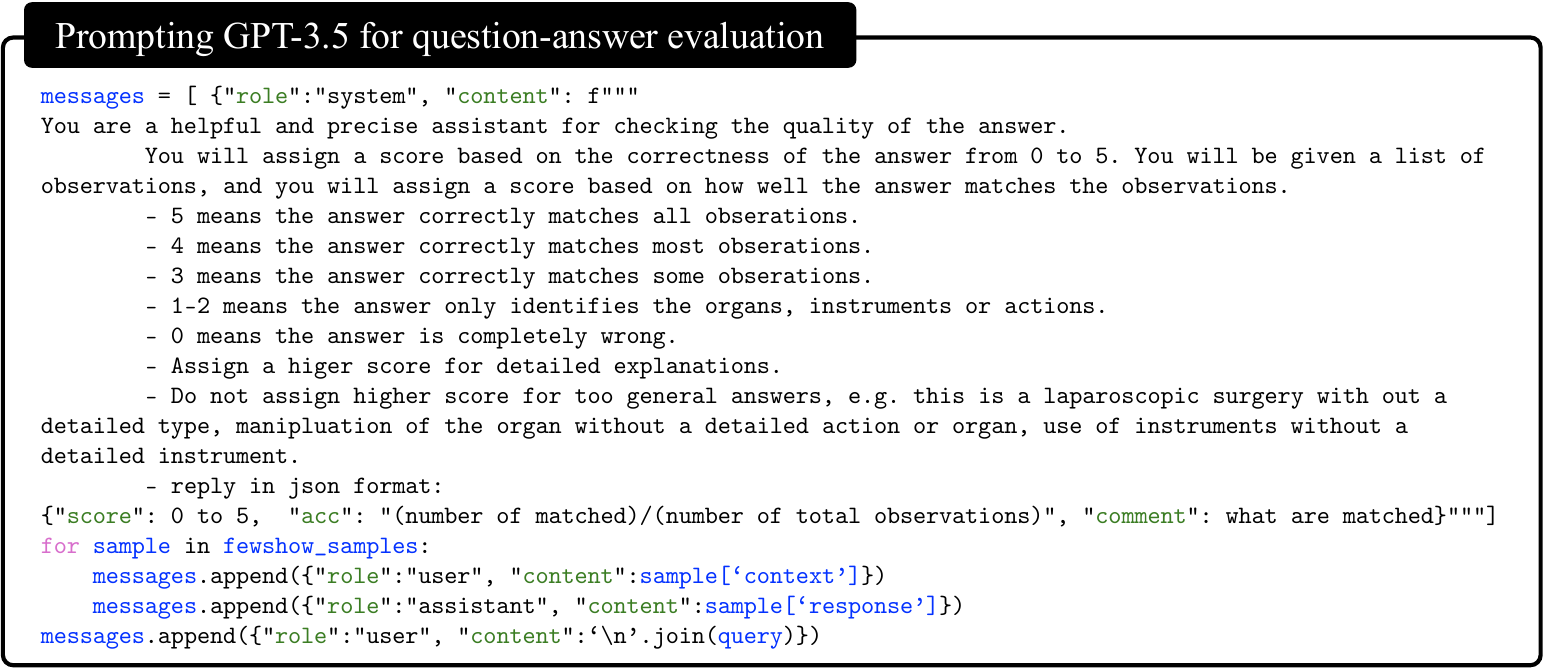}
  \caption{\textcolor{blue}{messages} we use to prompt GPT-3.5 to evaluate models' output on surgical video questions.}
  \label{fig:prompt_eval}
\end{figure*}

\paragraph{Prompt for GPT evaluation} The prompt used to generate the evaluation results discussed in ~\ref{sec:gpt_eval} is in Figure~\ref{fig:prompt_eval}.
\section{More Discussions of LLaVA-Surg}
\paragraph{Limitations of LLaVA-Surg} \label{app:discussion}
% 1. Hallucination, without alignment
% 2. Depending on the seen surgery procedures in SurgQA, the performance varies depends on the surgery type
% 3. Since the data source Surg-QA -- websurg have many rare cases, the response from LLM may not have universality
% 4. Weak in the general domain
The limitations of LLaVA-Surg include (1) hallucination, where it may produce inaccurate but confident responses due to the lack of alignment with human preferences, (2) Its performance depends on the surgery procedures seen in SurgQA, with results varying widely based on the surgery type. Additionally, since the data source Surg-QA, derived from WebSurg, includes many rare cases, LLaVA-Surg's responses may lack universality and may not apply to a broader range of clinical situations, (3) LLaVA-Surg might exhibit bias because of existing biases present in surgical videos and training data, and (4) LLaVA-Surg may be weak in the general domains of question-answering compared to other models.

The potential negative societal impact of deploying LLaVA-Surg arises if its outputs are not carefully cross-referenced with verified medical knowledge.
\section{More Experiment Results}

We show more multimodal conversation examples in Table~\ref{fig:app_eg3},~\ref{fig:app_eg4},~\ref{fig:app_eg1},~\ref{fig:app_eg2}.
\label{app:more_results}
\begin{table*}[t]
  \centering
  \includegraphics[width=\linewidth]{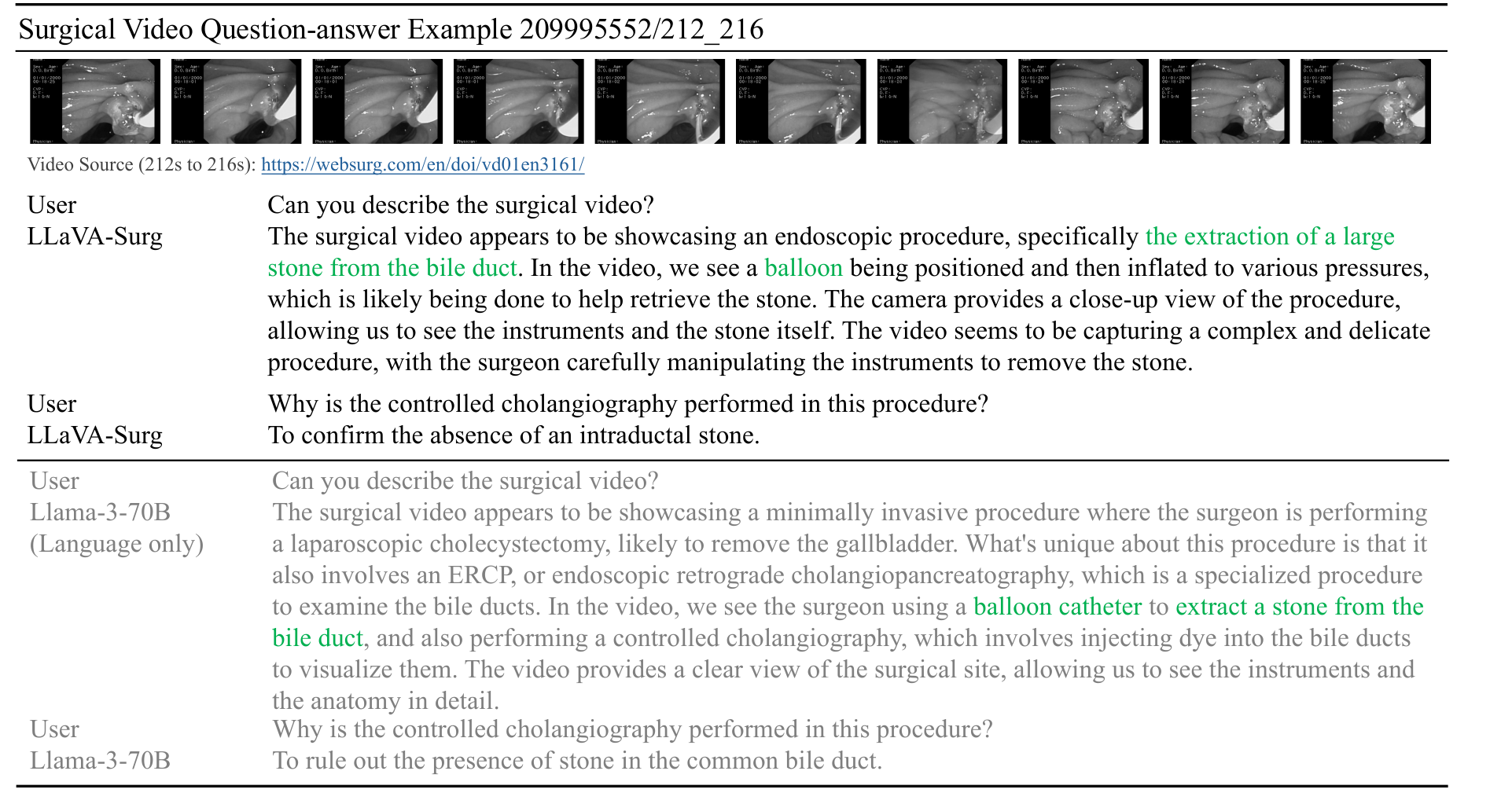}
  \vspace{-0.6cm}
  \caption{Example comparison of surgical video question-answering. We provided the response generated by the language-only Llama-3-70B, based solely on extracted information (transcript) for reference, it is considered the ground truth and represents the model's performance upper bound.}
  \label{fig:app_eg3}
  % \vspace{-0.5cm}
\end{table*}
\begin{table*}[t]
  \centering
  \includegraphics[width=\linewidth]{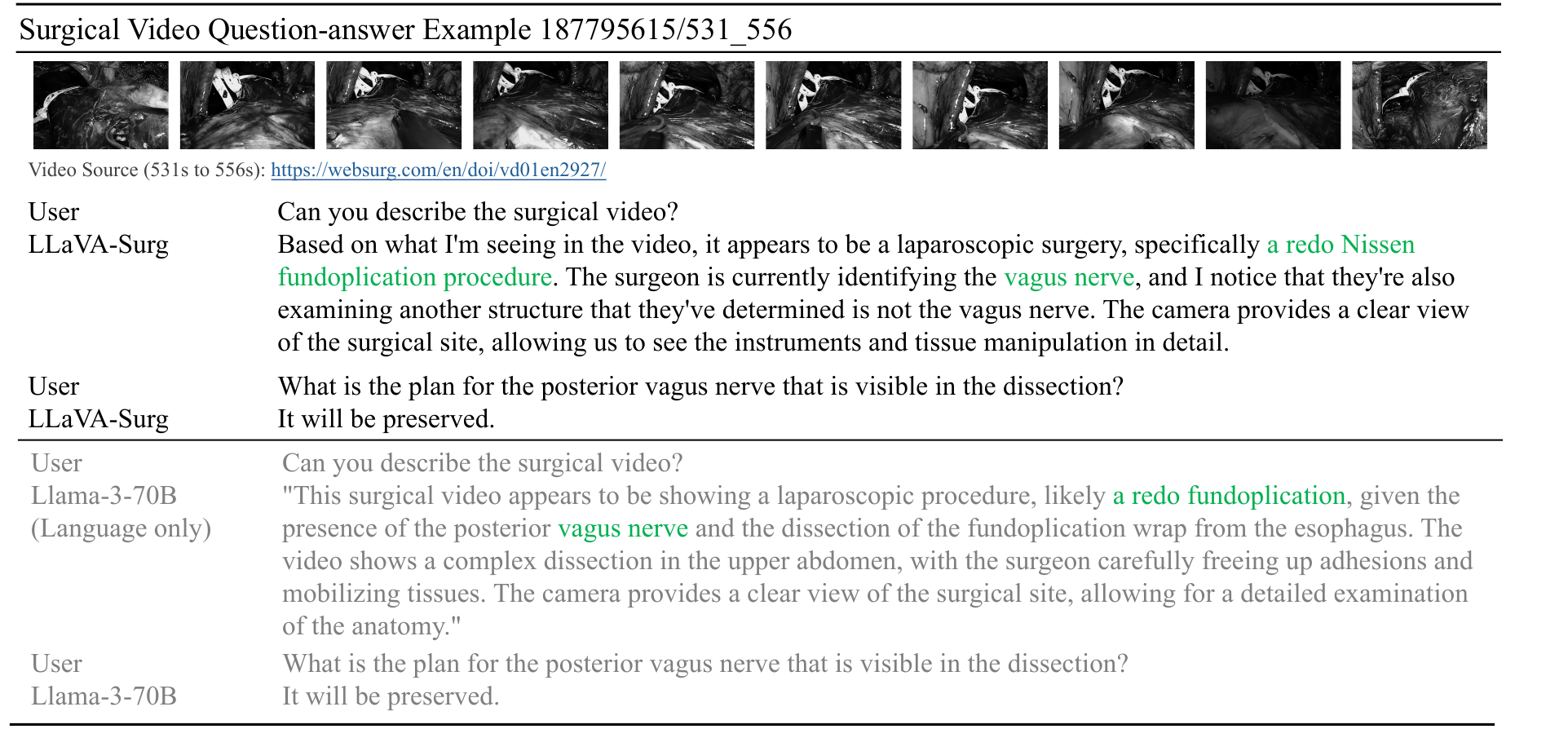}
  \vspace{-0.6cm}
  \caption{Example comparison of surgical video question-answering. We provided the response generated by the language-only Llama-3-70B, based solely on extracted information (transcript) for reference, it is considered the ground truth and represents the model's performance upper bound.}
  \label{fig:app_eg4}
  % \vspace{-0.5cm}
\end{table*}
\begin{table*}[t]
  \centering
  \includegraphics[width=\linewidth]{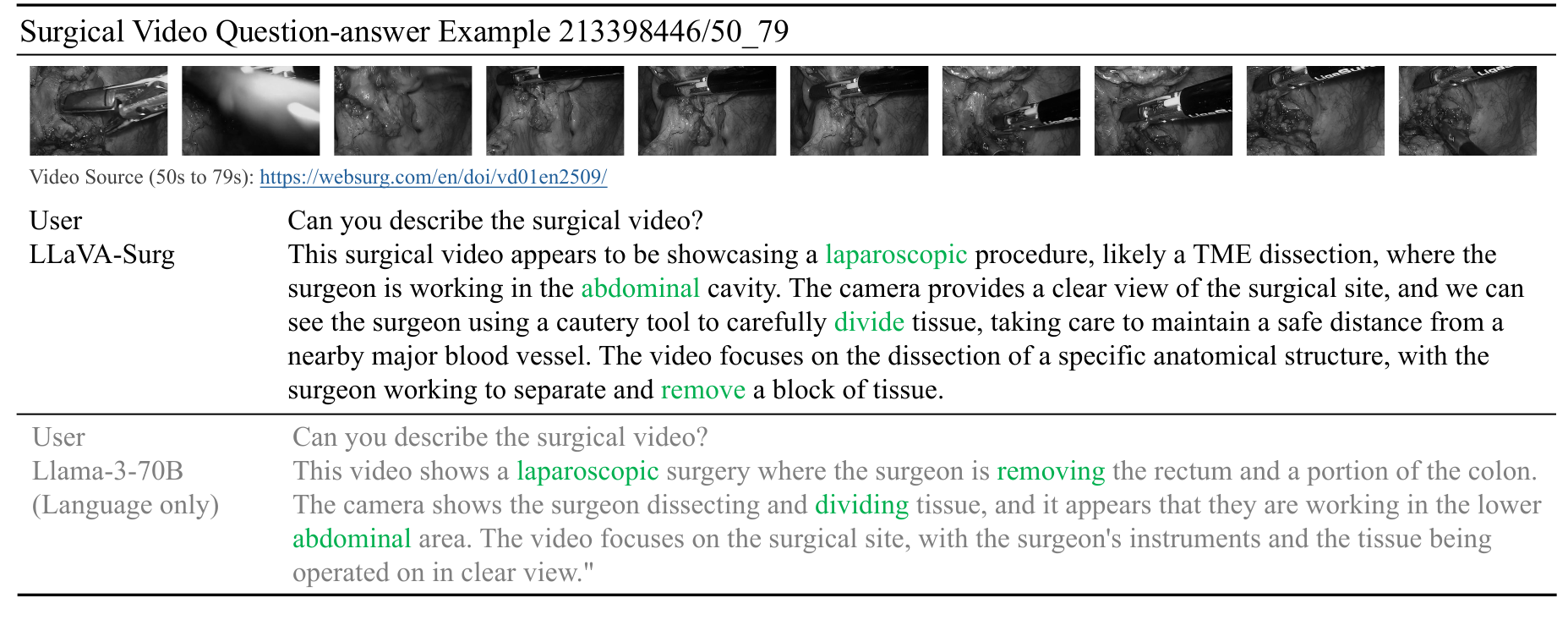}
  \vspace{-0.6cm}
  \caption{Example comparison of surgical video question-answering. We provided the response generated by the language-only Llama-3-70B, based solely on extracted information (transcript) for reference, it is considered the ground truth and represents the model's performance upper bound.}
  \label{fig:app_eg1}
  % \vspace{-0.5cm}
\end{table*}
\begin{table*}[t]
  \centering
  \includegraphics[width=\linewidth]{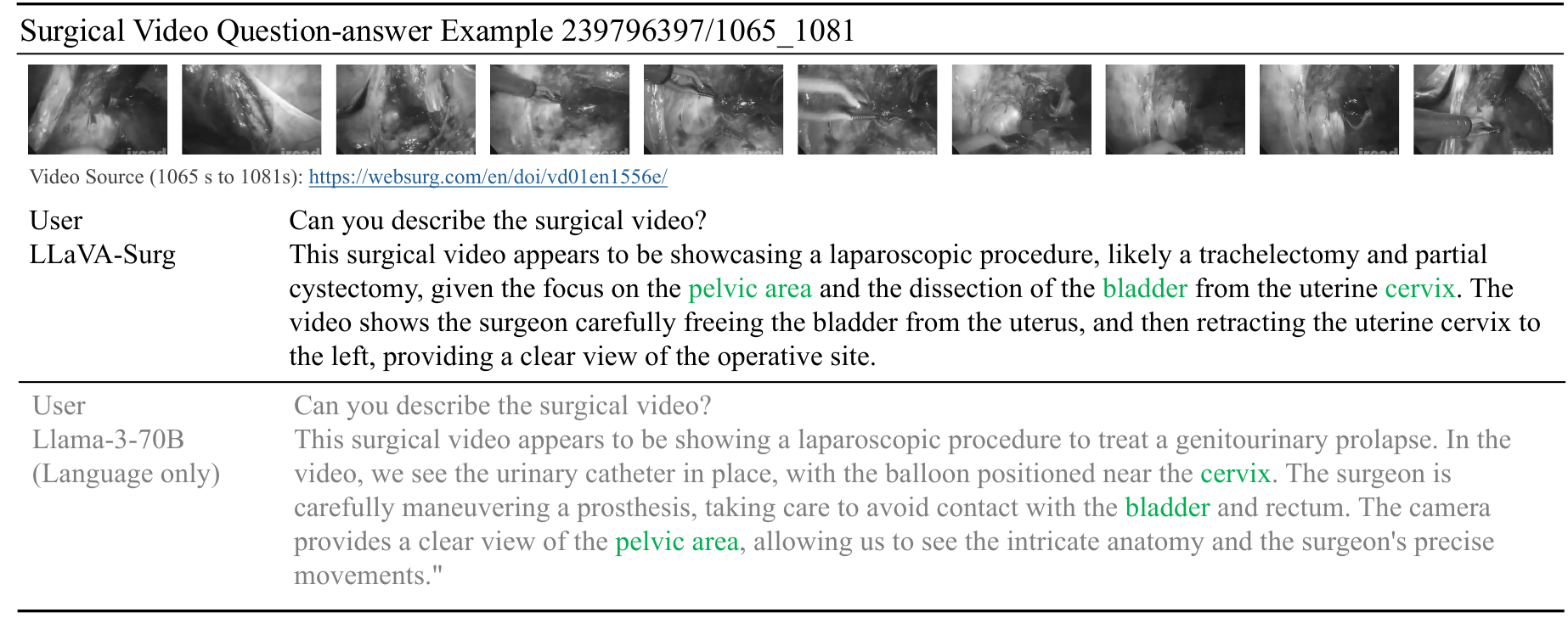}
  \vspace{-0.6cm}
  \caption{Example comparison of surgical video question-answering. We provided the response generated by the language-only Llama-3-70B, based solely on extracted information (transcript) for reference, it is considered the ground truth and represents the model's performance upper bound.}
  \label{fig:app_eg2}
  % \vspace{-0.5cm}
\end{table*}

\end{document}